\documentclass[10pt,twocolumn,letterpaper]{article}

\usepackage{cvpr}
\usepackage{times}
\usepackage{epsfig}
\usepackage{graphicx}
\usepackage{amsmath}
\usepackage{amssymb}

\usepackage{subcaption}
\usepackage{amsmath}
\usepackage{booktabs}
\usepackage{xcolor}
\usepackage{multirow}
\newcommand{\paragraphTitle}[1]{\vspace{2mm}\noindent{\textbf{#1}\hspace{2mm}}}
\newcommand{\Skip}[1]{}
\newcommand{\YOON}[1]{\textcolor{red}{\bfseries{YOON: {#1}}}}
\newcommand{\huo}[1]{{\textcolor[rgb]{0.0,0.3,0.8}{\emph{Huo: #1}}}}

\usepackage[breaklinks=true,bookmarks=false]{hyperref}

\cvprfinalcopy 

\def\cvprPaperID{5091} 

\ifcvprfinal\fi
\begin{document}

\title{Single Image Reflection Removal with Physically-Based Training Images}

\author{
	 Soomin Kim \hspace{13mm} Yuchi Huo \hspace{13mm} Sung-Eui Yoon\\
Korea Advanced Institute of Science and Technology (KAIST)\\
}
\maketitle
\thispagestyle{empty}

\begin{abstract}
	Recently, deep learning-based single image reflection
	separation methods have been exploited widely. To benefit the learning
	approach, a large number of training image-pairs (i.e., with and without
	reflections) were synthesized in various ways, yet they are away from a
	physically-based direction.	In this paper, physically based rendering is used for faithfully
	synthesizing the required training images, and a corresponding network
	structure and loss term are proposed. We utilize existing RGBD/RGB
images to estimate meshes, then physically simulate the light transportation
between meshes, glass, and lens with path tracing to synthesize training data,
which successfully reproduce the spatially variant anisotropic visual effect
of glass reflection. For guiding the separation better, we additionally
consider a module, backtrack network ($BT$-net) for backtracking the
reflections, which removes complicated ghosting, attenuation, blurred and
defocused effect of glass/lens. This enables obtaining a priori information
before having the distortion. The proposed method considering additional a
priori information with physically simulated training data is validated with
various real reflection images and shows visually pleasant and numerical
advantages compared with state-of-the-art techniques.
\end{abstract}

\section{Introduction}
\label{sec:intro}

When taking a photo through a glass or a window, 
the front scene that is transmitted through the glass can be seen, but sometimes
the reflection from the back scene is captured as well. These inevitable reflections
and dim transmission can be annoying for some cases, for example, a case of
taking a photo of a skyscraper from an indoor room.  As a result, removing the
reflections from the input images can help us to generate better images and
various computer vision techniques to work robustly. 

Physically, an image $I$ with those reflections is a sum of the glass
reflected back scene, $\tilde{R}$, and the glass transmitted 
front scene, $\tilde{T}$, as
$I(x,y) = \tilde{T}(x,y)+ \tilde{R}(x,y)$.
Single image reflection removal problem is ill-posed, without using additional assumptions or priors. 

\begin{figure}[t]
	\begin{center}
		\includegraphics[width=\linewidth]{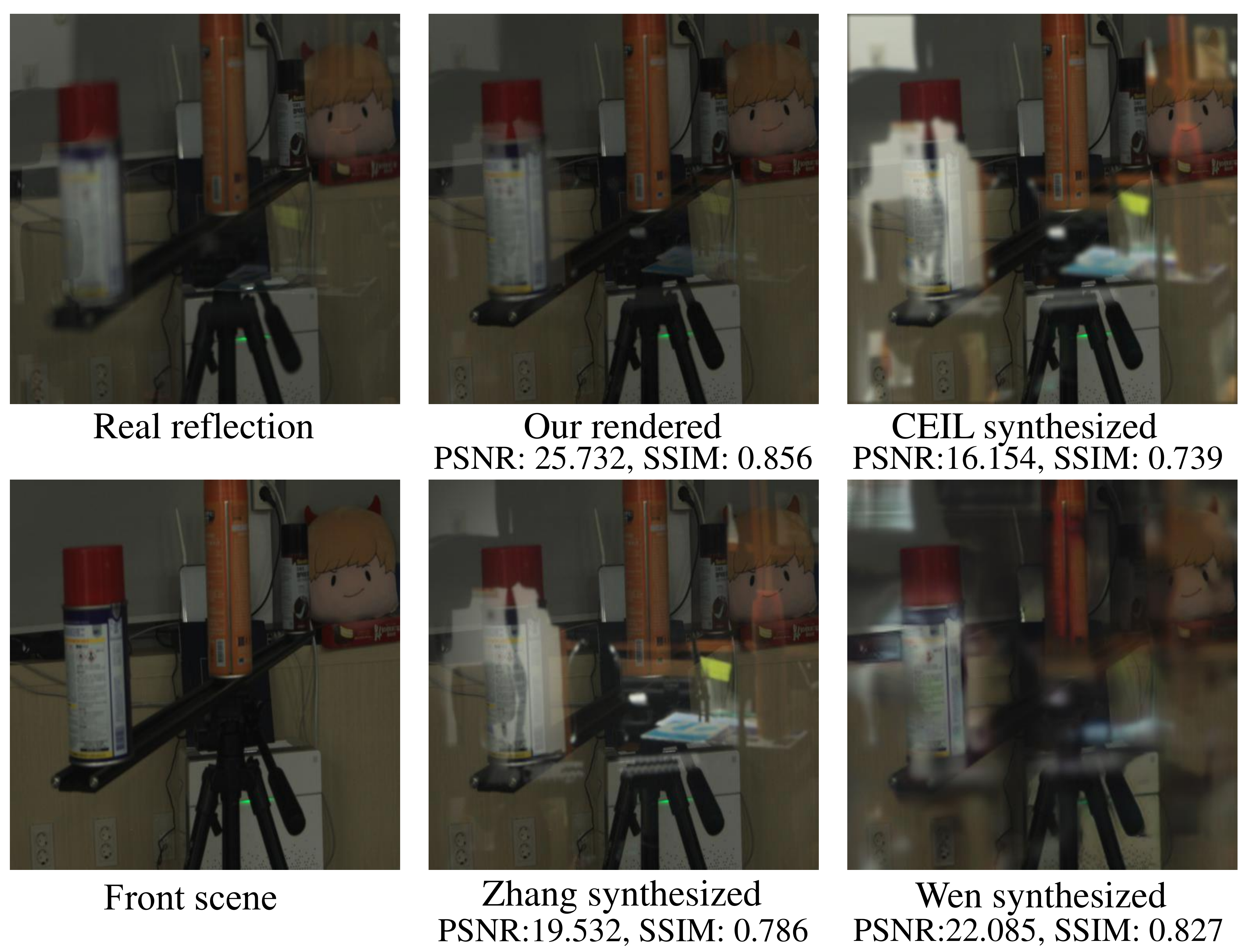}
	\end{center}
	\vspace{-5mm}
	\caption[]{\label{fig:datacomp}\small{
		Comparison between existing reflection synthesizing methods and
our physically-based rendering method. The real reflection image is captured by a camera behind a glass. Our method can produce spatially variant
visual effects that are most similar to the real-world reflection image. For
example, the near bottle is blurred and the far bottle is focused in the
transmission scene.  Also, the reflection level of some back scene objects is properly
	considered. In contrast, previous methods assume the glass transmitted front scene is all
	clear, and the reflected back scene is spatially invariantly blurred,
	introducing biased information to the dataset.
	Sec.~\ref{sec:DataAnalysis} has more details.}	}
\vspace{-4mm}
\end{figure}

Previous methods utilize multiple images of reflection with different conditions
for obtaining some
priors~\cite{agrawal2005removing,schechner2000separation,kong2014physically,punnappurath2019reflection}.
Especially, motion cue prior is widely used for
separating the reflections from multi-images~\cite{gai2012blind,xue2015computational,han2017reflection}.   
Although multiple-image reflection separation methods show reasonable results,
it is not easy for users to capture constrained images as suggested in the
prior approaches.

For single image reflection removal, natural image
priors~\cite{levin2007user,levin2003learning,levin2004separating} or smoothness
priors~\cite{li2014single,wan2016depth} are used for formulating
objective functions. Recent approaches started to utilize deep neural networks for removing the reflections on a single 
image. While training deep neural networks relies on a faithful dataset, most up-to-date
methods synthesize datasets in an image space through a weighted addition
between the front scene and the
back scene~\cite{fan2017generic,zhang2018single,wan2018crrn,yang2018seeing,wei2019single},
due to the difficulty of the physical simulation of the reflection and the
transmission phenomena. Recently, Wen et al.\cite{wen2019single} propose a
method that generates reflection training images using deep learning
architecture. However, these image-space methods ignore the physical fact that
the visual effects of reflections are spatially variant, depending on the 3D
positions of the visible points. Figure~\ref{fig:datacomp} shows the visual
and numerical comparison of generated reflection
images against the ground truth~(Sec.~\ref{sec:DataAnalysis}).

In this paper, we present a data generation method to synthesize physically faithful training data. The method is
based on modeling and rendering techniques,
such as depth estimation, geometry synthesizing, and physically-based rendering. We 
utilize such physically-based training images, including the transmission
and the reflection with or without glass/lens-effects, i.e., the attenuation,
defocusing, blurring, and ghosting effects related to passing through a
glass/camera lens (Sec. \ref{sec:data_generation}), for training our deep
learning architectures. Especially, we train a backtrack network ($BT$-net) to
fetch a priori information to improve separation quality.
 
In summary, our contributions are as follows:
\begin{itemize}
	\item Propose a synthesizing method to physically render a faithful
reflection image dataset for training.
	\item Use $BT$-net to transform the reflection image back to its prior-distortion status as a priori information of the separation problem.	

\end{itemize}

\section{Related Work}
\label{sec:relatedwork}

\begin{figure*}[t]
	\begin{center}
		\includegraphics[width=0.8\linewidth]{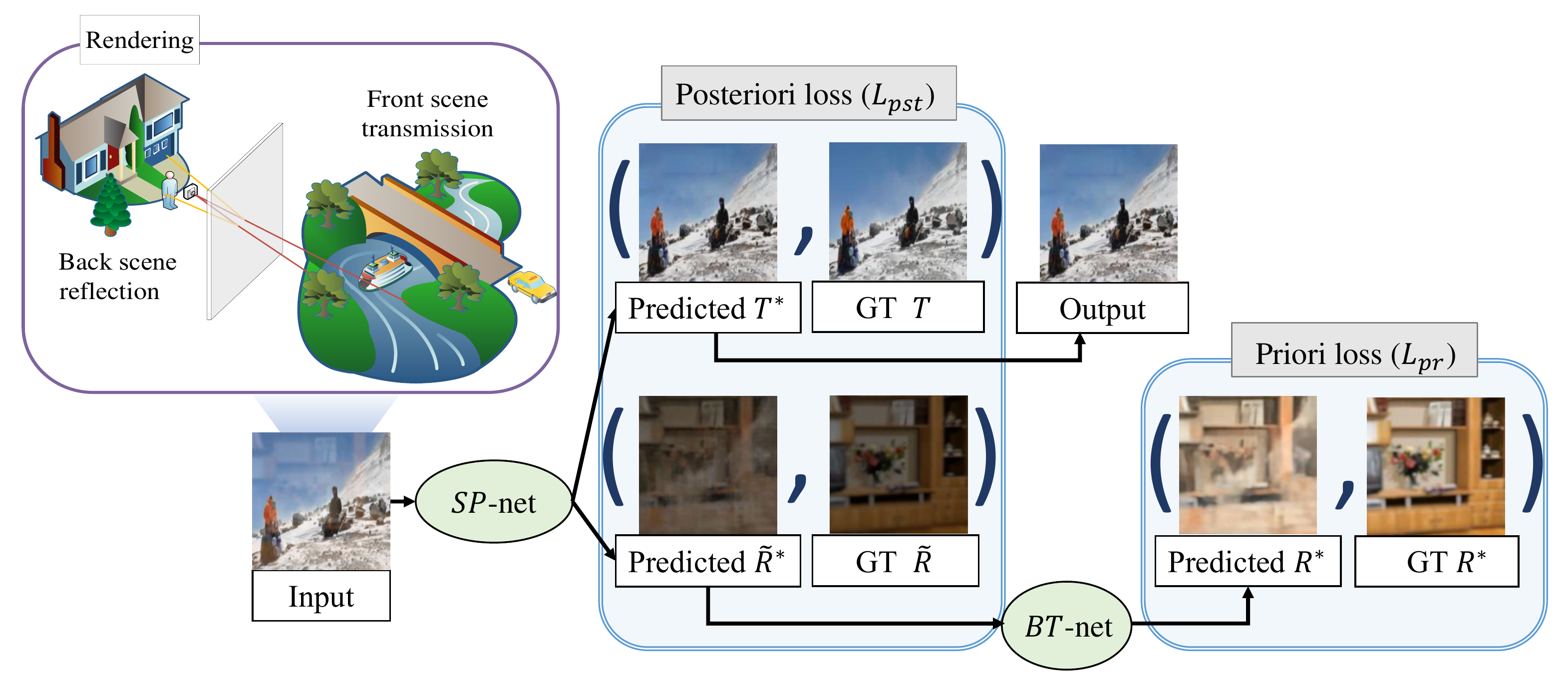}
	\end{center}
	\caption[]{\label{fig:alg_diagram}
		\small{
			Overview of our method structure. From a given image
with reflection ($I$), our $SP$-net first separates $I$ to predicted front scene
transmission, $T^*$, and back scene reflection with glass-effects,
$\tilde{R}^*$.
A posteriori loss ($L_{pst}$) is calculated with each
of the predicted values and its ground truth. 
Our trained backtrack network,
$BT$-net, removes the glass and lens effects of the predicted
 $\tilde{R}^*$ into $R^*$. Since  $R^*$ is released from complicated glass/lens-effects, we can
better capture various image information, resulting in clearer error matching 
between the predicted image and its ground truth. To utilize this
information, we use a new loss, a priori loss ($L_{pr}$), between $R^*$
and its ground truth (GT). The entire separation network is trained with a loss
combination of $L_{pst}$ and
$L_{pr}$. 
	} 	
	}	
\end{figure*}

\paragraphTitle{Single image-based methods with conventional priors.}
Since the single image
methods lack information compared to the multi-image methods, they assume
predefined priors. One of the widely used priors is the natural image gradient
sparsity priors~\cite{levin2003learning,levin2004separating}. These approaches
decompose the layers with minimal gradients and local features. Levin et al.~\cite{levin2007user}
proposed gradient sparsity priors with user
labeling and showed reasonable results. 
Another widely used assumption is that reflection layers are more likely to be blurred because of
the different distance to the camera~\cite{li2014single,wan2016depth}.
In addition to that, Arvanitopoulos et al.~\cite{arvanitopoulos2017single} proposed the Laplacian fidelity term and $l_0$-gradient sparsity term to suppress reflections. Shih et al.~\cite{shih2015reflection} suggested to examine ghosting effects on the reflection and model them by Gaussian Mixture Models (GMM) patch prior.

\paragraphTitle{Single image based methods with deep learning.} 
Recent studies start to adopt deep learning for
the reflection removal problem. Fan et al.~\cite{fan2017generic} proposed
a two-step deep architecture utilizing edges of the image.
Zhang et al.~\cite{zhang2018single}
adopt conditional
GAN~\cite{isola2017image} with a combination of perceptual loss, adversarial
loss, and exclusion loss for separating reflection. Wan et
al.~\cite{wan2018crrn} suggested a concurrent deep learning-based framework for
gradient inference and image inference. Yang et al.~\cite{yang2018seeing} proposed a
cascaded deep network for estimating both the transmission and the
reflection. Wei et al.~\cite{wei2019single} suggest to
use misaligned real images for training and its corresponding loss term, and
Wen et al.~\cite{wen2019single} proposes a learning architecture to
produce reflection training images with a corresponding removal network.

Our method is also based on learning-based single image reflection removal, but
with two main differentiations. First, we render a physically faithful
dataset to reproduce lens focus and glass-effect realistically. These spatially
variant anisotropic visual effects vary depending on the depth and viewing
angle across the image space and were not supported faithfully by previous
image-space data generation methods.  Second, our method utilizes information
not only after the images were distorted by the glass/lens (a posteriori
information), but also before the glass/lens distortion (a priori information),
to get better separation results.

\paragraphTitle{Synthesizing training datasets with rendering.}
Monte Carlo (MC) rendering is widely used in various applications for
high-quality image synthesis. Its theoretical foundation includes the physical
simulation of light transportation and the unbiased integration of incident
radiances~\cite{Yoon18}. In order to simulate the shading effect of complex
geometry details, displacement mapping is proposed to reconstruct geometry from
a depth map \cite{donnelly2005per}. Because physically-based rendering can
faithfully simulate the physical process of light transportation, it has been
proven to be a promising way to synthesize deep learning datasets for various
computer vision problems~\cite{zhang2017physically,Su_2015_ICCV,rozantsev2015rendering}.

In this paper, we propose to use displacement mapping and path tracing to
synthesize a physically plausible dataset for the reflection removal problem.

\section{Overview}
\label{sec:overview}

In this section, we present an overview of our method.
There are two main components of our reflection removal technique. The first
part is synthetically generating training images with physically-based
rendering, and the second part is network training using the rendered training images as additional priori information.

To train the reflection removal network, a large
amount of reflection and reflection-free image pairs are necessary. It is,
however, quite troublesome to obtain such kinds of many image pairs.
Most of the prior deep learning-based reflection removal
methods~\cite{fan2017generic,zhang2018single,wan2018crrn,yang2018seeing,wei2019single}
synthesize a reflection image by mixing two ordinary images, one as a reflection
and another as a transmission, with different coefficients followed by applying
Gaussian blurring and scaling down the brightness of the reflection. The technical details vary from one to the other, but they synthesize the reflection images in image space. Lately, Wen et al.~\cite{wen2019single} suggested to use a network for generating reflection training image pairs, but still, they do not consider the spatially variant visual effects.

We find that instead of synthesizing the reflection images in the image space,
rendering the reflection images in a 3D space would produce
more realistic images for training, resulting in a higher removal accuracy. In
order to achieve a physically faithful dataset, we adopt a series of modeling and
rendering techniques, i.e., depth estimation, geometry synthesizing, and
physically-based rendering technique (path tracing~\cite{kajiya1986rendering}).

From existing DIODE RGBD dataset~\cite{diode_dataset} and PLACES365 RGB
dataset~\cite{zhou2017places}, we randomly choose one image as a
\textbf{front scene} transmission layer (\textit{the side in front of the
camera)} and another image as
a \textbf{back scene} reflection layer (\emph{the side behind the camera)}. 
With one front scene and one back scene as a \textbf{scene} setup, we extract
the 3D model of the scene with depth and then render it with path tracing to
synthesize a group of images with or without reflection for training; for the
RGB dataset, we apply depth estimation~\cite{chen2016single} to extract the 3D
model of the scene.

\begin{figure*}
	\centering
	\includegraphics[width=0.95\linewidth]{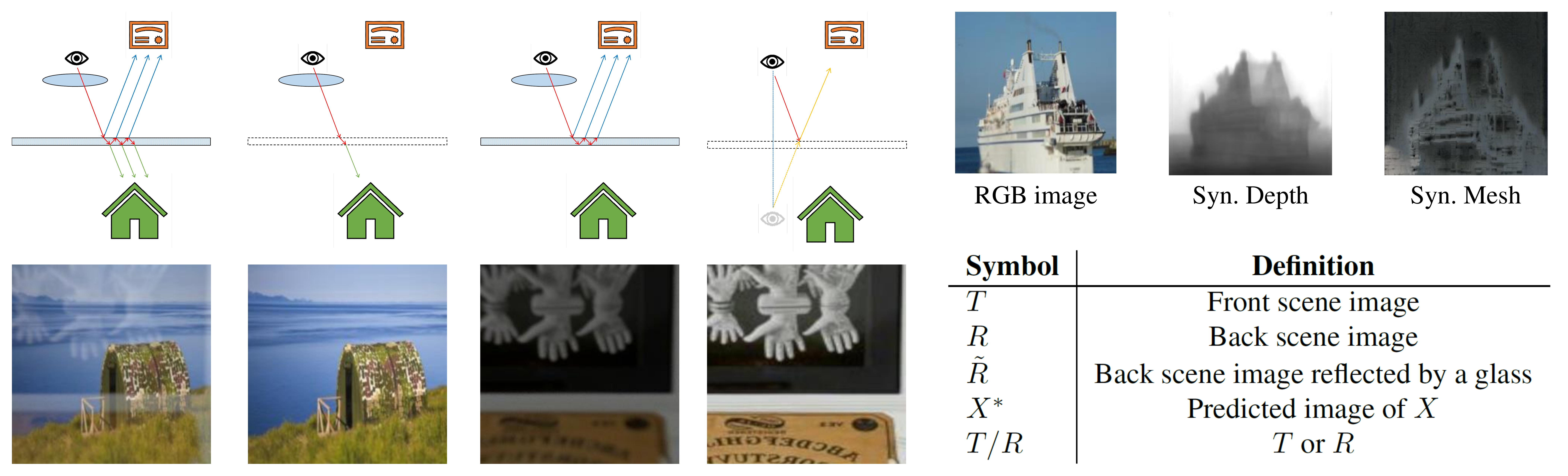}
	
	(1)~$I$\hspace{42pt} (2)~$T$\hspace{46pt}(3)~$\tilde{R}$\hspace{46pt}(4)~$R$\hspace{190pt}\vspace{7pt}
	\caption{\small{
		In this example, we set up a scene consisting of the front scene containing the
		house and back scene with indoor decorations. Suppose that we look at the front scene  with a camera behind a glass. (1) $I$ is the input
		image with reflection.
		(2) $T$ is front scene transmission.	
		(3) $\tilde{R}$ is the reflected back scene (reflection) image with lens/glass-effects, and it is computed by physically simulating the real-world attenuation and glass-effect, i.e., multiple bounces within the glass. 
		(4) $R$ is
		the back scene (reflection) image without any glass-effects.
	}}
	\label{fig2:dataset}
\end{figure*}

Figure~\ref{fig:alg_diagram} shows the overall pipeline of our network training
algorithm using 4-image tuples as the training ground truth (GT). The algorithm
contains a separation network ($SP$-net), which separates the input image into
two layers with the help of the backtrack network ($BT$-net), which attempts to
remove the glass/lens-effects (e.g., blurring, attenuation, and ghosting) of $\tilde{R}$ for better separation.

As shown in Figure~\ref{fig2:dataset}, we can render 4-image tuples ($I$,
$T$,$\tilde{R}$, $R$), and with those image tuples, we first train the
$BT$-net, so that the $\tilde{R}$ can be backtracked into $R$ and can be used
for additional a priori information for separation. The table of
Figure~\ref{fig2:dataset} summarizes these notations. 
We then train the main $SP$-net with rendered 4 tuples along with the pretrained $BT$-net.

Intuitively, the algorithm makes use of additional a priori information (without glass/lens-effects) of separated $\tilde{R}$ along with widely used a posteriori information with glass/lens-effects.
Specifically, those existing techniques try to calculate the error of separated reflection distorted by the glass-effect. 
However, the complicated glass-effects can hinder clear matching between predicted
images and their GTs (e.g., feature loss), resulting in a low-quality loss generation.
Interestingly, we find that the a priori information can provide
additional clues for the separation problem.
With the help of our $BT$-net, we can physically backtrack the physical process and remove the glass/lens-effects on an image.

\Skip{
\begin{table}[t]
	\begin{center}
		\caption{\small{Symbols with $\ast$ represent images predicted by the
				network; ones w/o $\ast$ represent rendered or captured
				ground-truth (GT) images.
		}}
		\label{tab1:symbols}
		\begin{tabular}{l|c r} 
			\textbf{Symbol} & \textbf{Definition} \\
			\hline
			$T$ & Front scene image \\
			$R$ & Back scene image \\
			$\tilde{R}$ & Back scene image reflected by a glass \\
			$ X^{\ast}$ & Predicted image of $X$ \\
			$T/R$ & $T$ or $R$  \\
		\end{tabular}
	\end{center}
	\vspace{-4mm}
\end{table}
}

\section{Physically Faithful Dataset Generation}\label{sec:data_generation}
\label{sec:dataset}
Compared with the classic image-space synthesized data, our physically faithful
data is featured with anisotropic spatial variations that rely on physical
simulation of light
transportation within 3D space. 
In theory, the glass-effect and its physical light transmission effect are much
more complex compared to the existing Gaussian blurring assumption adopted in
prior techniques~\cite{fan2017generic,zhang2018single,yang2018seeing}. For a light path 
connecting a visible point $\mathbf{x}_k$ and the camera viewpoint $\mathbf{x}_0$ (Figure~\ref{fig2:dataset}) bouncing through $k-1$ points, the contribution is computed as:

\begin{equation}
\begin{aligned}
L(\mathbf{x}_0&\leftarrow\mathbf{x}_k)=\frac{L_e(\mathbf{x}_k,\mathbf{x}_{k-1})\hat{V}(\mathbf{x}_{k-1},\mathbf{x}_k)}{prob(\mathbf{x}_0,\mathbf{x}_1,...\mathbf{x}_k)}\\
&\prod^{k-1}_{i=1}{G(\mathbf{x}_{i},\mathbf{x}_{i+1})f(\mathbf{x}_{i-1},\mathbf{x}_{i},\mathbf{x}_{i+1})}\hat{V}(\mathbf{x}_{i-1},\mathbf{x}_i),
\label{eq:rendering}
\end{aligned}
\end{equation}
where $L_e(\mathbf{x}_k,\mathbf{x}_{k-1})$ is the outgoing radiance of point
$\mathbf{x}_{k}$, $prob(\mathbf{x}_0,\mathbf{x}_1,...\mathbf{x}_k)$ is the
probability of sampling the path $\mathbf{x}_0,\mathbf{x}_1,...\mathbf{x}_k$
from a given sampler, $\hat{V}(\mathbf{x}_{i-1},\mathbf{x}_i)$ is the
generalized visibility term between two points considering the medium
attenuation factor, $G(\mathbf{x}_{i},\mathbf{x}_{i+1})$ is the geometry term
between
two points, and $f(\mathbf{x}_{i-1},\mathbf{x}_{i},\mathbf{x}_{i+1})$ is the
bidirectional scattering function of point $\mathbf{x}_i$ from
$\mathbf{x}_{i-1}$ to $\mathbf{x}_{i+1}$. Detailed explanations of these terms
can be found in 
\cite{dachsbacher2014scalable}. 

\begin{figure*}
	\centering
	\begin{subfigure}{.135\textwidth}
		\centering
		\includegraphics[width=.95\linewidth]{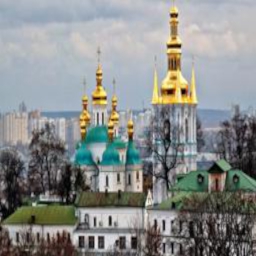}
		\caption{}
	\end{subfigure}
	\begin{subfigure}{.135\textwidth}
		\centering
		\includegraphics[width=.95\linewidth]{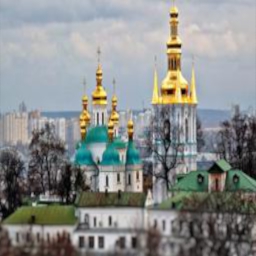}
		\caption{}
	\end{subfigure}
	\begin{subfigure}{.135\textwidth}
		\centering
		\includegraphics[width=.95\linewidth]{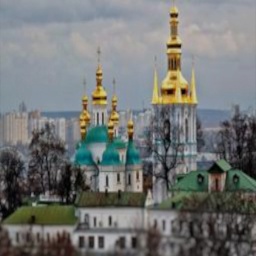}
		\caption{}
	\end{subfigure}
	\begin{subfigure}{.135\textwidth}
		\centering
		\includegraphics[width=.95\linewidth]{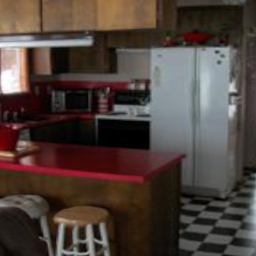}
		\caption{}
	\end{subfigure}
	\begin{subfigure}{.135\textwidth}
		\centering
		\includegraphics[width=.95\linewidth]{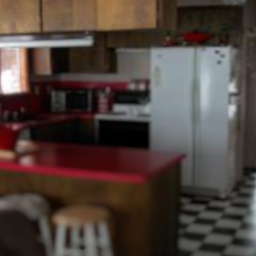}
		\caption{}
	\end{subfigure}
	\begin{subfigure}{.135\textwidth}
		\centering
		\includegraphics[width=.95\linewidth]{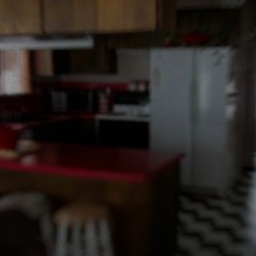}
		\caption{}
	\end{subfigure}
	\begin{subfigure}{.135\textwidth}
		\centering
		\includegraphics[width=.95\linewidth]{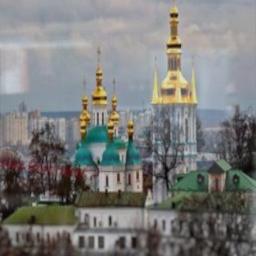}
		\caption{}
	\end{subfigure}
	\caption{
		\small{
		Images w/ and w/o lens- and glass-effects. (a) is a front scene w/o
		lens- and glass-effects; thus the whole image is sharp and clear. (b) is front scene
		w/ lens-effect, but w/o a glass-effect, where the corners are blurred since they are
		outside the focus range; the focus point is set to the center of the front scene and thus its effect is subtle. (c)
		is front scene w/ lens- and glass-effect, where the color is attenuated and image are
		even more blurred due to the glass. (d) is a back scene w/o lens- and glass-effects,
		so it is clean. (e) is back scene w/
		lens-effect, but w/o glass-effect, where the whole image is blurred.
		(f) is a front scene w/ lens- and glass-effects, where the glass further introduces
		attenuation, blurring and ghosting effect. (g) is the sum of (c) and (f).}
	}
	\label{fig:render}
\end{figure*}

Simply speaking, a light path starting from a visible point is
reflected/refracted multiple times by the glass and lens before contributing
its brightness to the image, resulting in ghosting, blurring, defocusing, and
attenuation. We call the visual effects resulting from passing through the lens
or glass as lens/glass-effects. Lens-effect includes defocusing and
attenuation. Glass-effect includes ghosting, blurring, and attenuation. 
When a path segment between $\mathbf{x}_i$ and $\mathbf{x}_{i+1}$ passes
through glass/lens, it will introduce glass/lens-effects. To
remove those effects, we can render a scene without a glass or lens (Figure~\ref{fig:render}).

All these visual effects are spatially variant because the contribution
function (Equation~\ref{eq:rendering}) is defined in 3D space rather 2D image
space. In order to prepare such a dataset, we adopt a series of modeling and
rendering techniques.
Our physically-synthesized dataset not only
improves the network performance but also provides a new perspective for
understanding and exploring the reflection removal problem
based on a physical ground.

\subsection{Mesh Generation}

Generating a variety of geometry meshes is the first block enabling physical
simulation. Because modeling thousands of geometry scenes is economically
prohibitive, we adapt the existing DIODE RGBD dataset~\cite{diode_dataset}. In order
to expand the diversity of the dataset, e.g., to add scenes with humans, we additionally use the labeled RGB dataset for scene
recognition~\cite{zhou2017places} and adopt a depth estimation
technique~\cite{chen2016single} to synthesize the depth channel. 

We choose 3000
image pairs (6~k in total) from the DIODE dataset, and 2000 image pairs (4~k in total) from
the PLACES dataset. Specifically, we selected 34 categories of the scenes from
the PLACES dataset. Because the depth estimation method predicts only normalized
relative depth in a single image, we manually scaled each category of the scene
with an
appropriated depth range; e.g., 4~m depth on average for the bedroom scene.
We mix 3000 scanned RGBD image pairs and 2000 synthesized RGBD pairs. \Skip{\huo{The synthesized RGBD image pairs might introduce certain errors in depth distribution. However, we found that the benefit exceeds the harm as discussed in} \huo{Now we need to remove this sentence but it might be a weak point to be attacked.}}
Finally, the depth channel is fed into Blender \cite{Blender} as a displacement map to export a geometry mesh from the input image. The  figures in the top right corner of Figure~\ref{fig2:dataset} show an example.

\subsection{Rendering Process}
\label{sec4.2:rendering_process}

Given an RGB image and its corresponding mesh geometry, we attach the RGB
channels of the image to the geometry surface to simulate the physical light transportation with path tracing~\cite{Mitsuba}. For each scene setup, we randomly
choose two images out of our image dataset, one for the front scene and the other for the back scene and render the entire scene with a glass model in the middle. 

We study and decompose the physical process of light transportation and to fetch a posteriori and a priori information by rendering up to 
four different images for each scene.
Figure \ref{fig2:dataset} shows the illustrations of these four different images for a scene.  
These four different rendered images include: 

\begin{itemize}
	\item $I$: An input image containing transmission plus reflection,
where both front scene and back scene are rendered with the glass-effect and lens-effect.
\Skip{
\item $\tilde{T}$: The front scene image transmitted by glass with glass-effect. 
}
\item $T$:
	The front scene image without any glass-effect. We simulate it with a
\textbf{virtual glass}, instead of the real glass, that warps the light path as real glass, but does not cause any ghosting, blurring, and attenuation effect. 
\item $\tilde{R}$: The back scene image reflected by a glass with glass- and lens-effects.
\item $R$: The back scene reflection image without any glass-effect and lens-effects. We
	simulate it also with the \textbf{virtual glass} to calculate the
		reflective direction.

\end{itemize}

Note that exact $T$ and $R$ are
actually impossible to be captured by the real camera because taking away a
real-world glass will certainly make image points shifted and thus misaligned
with $I$ anymore.  

All images are rendered with a low-discrepancy sampler~\cite{Mitsuba} 
with 256 samples per pixel, which is large enough to restrain visible noises. 
The glass is $10$ millimeters of thickness with a common refractive index of $1.6$, 
placed $30$ centimeters in front of the camera. We use $55$
millimeter thin lens model with a focus radius of $0.00893$. In order to simulate
the real application scenario, we set the focus distance to the center of the
front scene. Overall, our synthetically generated dataset has 5000 image tuples for training and 200 image tuples for testing.  

\Skip{
Beside the rendered dataset, we do not use any real-world images
for training because the proposed network is focused on 5-image tuples in a
component-wise manner. 


This is mainly because we found that training with
images from a real-world dataset, but testing with another real-world dataset
did not achieve the best quality due to the domaing gap. On the other hand,
using a same dataset in training and testing significantly embellishes the
results compared with general cases\YOON{cannot follow the logic}.
}

\section{Proposed Network Architectures}
\label{sec:proposed_method}

Our model consists of two sub-networks. As illustrated in Figure
\ref{fig:alg_diagram}, there is a backtrack network for the back scene reflection
($BT$-net) and a main separation network ($SP$-net). Initially, the input image
$I$ is separated into ${T}^{\ast} $ and $\tilde{R}^{\ast}$(with glass-effect)
using the $SP$-net, and then $\tilde{R}^{\ast}$ is fed into $BT$-net for
removing the glass/lens-effect such as distortion, ghosting, attenuation, and
defocusing.  The output of $BT$-net is $R^{\ast}$, which is supposed to be
devoid of the glass/lens-effect, and is  used for providing additional error
calculation for $SP$-net (a priori loss). 
Each of our network input is concatenated with  
extracted hypercolumn features~\cite{hariharan2015hypercolumns} from the VGG-19
network~\cite{simonyan2015very} as an augmented input for
better utilizing semantic information~\cite{zhang2018single}.

\subsection{Loss function}
\label{sec:5.2_loss}

Each sub-network has three loss terms: \textbf{$l_1$-loss, feature
loss,} and \textbf{adversarial loss}. $l_1$-loss is used for penalizing
pixel-wise difference in the predicted one, say, $X^{\ast}$, and its
GT, $X$, via $l_1 = \|X^{\ast} - X\|$ for low-level information
comparison for the results.
Our feature loss and adversarial loss are based on \cite{zhang2018single}. 
The feature loss $L_{ft}$ (Eq.~\ref{equ:feature_loss}) is used for
considering semantic information, based on the activation difference of a
pre-trained VGG-19 network $\Phi$, which is a trained with the ImageNet
dataset~\cite{russakovsky2015imagenet}.  For obtaining realistic images, the
adversarial loss is adopted too, as the
many other recent studies~\cite{zhang2018single,yang2018seeing,ledig2017photo,zhu2017unpaired}.
A conditional GAN~\cite{isola2017image} is utilized for this. 
For explanation, suppose that one of our sub-network's generator is $f$, its
input is $X$, and its GT is $Y$.  The feature loss $L_{ft}$ is
calculated as follows:
\begin{equation}
\begin{aligned}
L_{ft}(f(X), Y) = \sum_{l}^{} \gamma \|\Phi_l(Y) - \Phi_l(f(X))\|,
\label{equ:feature_loss}
\end{aligned}
\end{equation} 
\noindent where $\Phi_l$ indicates the $l$-th layer of the VGG-19 network with
the same layer selection of \cite{zhang2018single}, which is `conv1\textunderscore2',  `conv2\textunderscore2', `conv3\textunderscore2', `conv4\textunderscore2', and `conv5\textunderscore2'. $\gamma$ is the weighting
parameter, which is empirically set to $0.2$. 

For the adversarial loss, the discriminator $D$ of one sub-network is trained
by:
  \begin{equation}
\begin{aligned}
 \sum_{X, Y \in \mathcal{D}}^{} \log D(X,f(X)) - \log D(X, Y),
\label{equ:discriminator_loss}
\end{aligned}
\end{equation} 
where the discriminator tries to differentiate between the GT patches of $Y$
and patches given by $f(X)$ conditioned on the input $X$. Adversarial
loss is then defined as follows:
  \begin{equation}
\begin{aligned}
L_{adv}(X, f(X)) = \sum_{X \in \mathcal{D}}^{}- \log D(X,f(X)).
\label{equ:adv_loss}
\end{aligned}
\end{equation} 

\paragraphTitle{Loss for $SP$-net.}
The purpose of the $SP$-net is separating $T^*$ and $\tilde{R}^*$ from
the input $I$.  
The first loss we calculate for training $SP$-net on its output (${T}^*,
\tilde{R}^*$) is \textit{a posteriori loss} ($L_{pst}$) with the lens/glass-effect. It is the
combination of $l_1$-loss, feature loss between predicted value and ground-truth,
and adversarial loss for ${T}^*$.  After using $BT$-nets removing
glass/lens-effect of $\tilde{R}^*$ (so that it becomes ${R}^*$), we also calculate the second loss
term called \textit{a priori loss} ($L_{pr}$) without the glass/lens-effect between predicted $ R^*$
and ground-truth $R$.
\vspace{-3mm}
\begin{subequations}
\begin{align}
L_{pst} &=L_{l_1}({T}^*,{T})+  L_{ft}({T}^*, {T}) + L_{adv}(I, {T}^*) \notag\\
        &+ L_{l_1}(\tilde{R}^*,\tilde{R}) + L_{ft}(\tilde{R}^*, \tilde{R}),  \\
\begin{split}
L_{pr} &= L_{l_1}(R^*, R)+  L_{ft}(R^*,R)  . 
\end{split}
\label{equ:SPnet_loss}
\end{align}
\end{subequations}

Combining the above loss terms, our complete loss for $SP$-net is
$L_{SP} = L_{pst} + L_{pr}$.

\paragraphTitle{Loss for $BT$-net.}
The goal of $BT$-net is removing the glass/lens
effects from $\tilde{R}$, so that it can be recovered from darkening and blurring. To train the network, we formulate a combined loss function of $l_1$-loss, feature loss, and adversarial loss as follows:
\begin{flalign}
L_{BT} = L_{l_1}(R^*,R)+  L_{ft}(R^*, R)+ L_{adv}(\tilde{R}, R^*).
\label{equ:GRnet_loss}
\end{flalign} 

\Skip{ 
\begin{figure}[b]
	\centering
	\begin{subfigure}{.088\textwidth}
		\centering
		\includegraphics[width=\linewidth]{images/114-input.jpg}
		\caption*{Input}
	\end{subfigure}\hspace{4mm}
	\begin{subfigure}{.088\textwidth}
		\centering
		\includegraphics[width=\linewidth]{images/114-wo_GR.jpg}
		\caption*{Result of (1)}
	\end{subfigure}\hspace{4mm}
	\begin{subfigure}{.088\textwidth}
		\centering
		\includegraphics[width=.95\linewidth]{images/114-wo_priori.jpg}
		\caption*{Result of (2)}
	\end{subfigure}\hspace{4mm}
	\begin{subfigure}{.088\textwidth}
		\centering
		\includegraphics[width=.95\linewidth]{images/114-complete.jpg}
		\caption*{Result of (3)}
	\end{subfigure}
	\caption{
		Visual results of our ablation study. Our complete
	model (3) removes more of the reflection compared to the ablated models (1) and (2) shown in Table~\ref{tb1:ablation}. 
}
	\label{fig:ablation_example}
\end{figure}

}

\begin{figure}[h]
	\centering
	\begin{subfigure}{.11\textwidth}
		\centering
		\includegraphics[width=\linewidth]{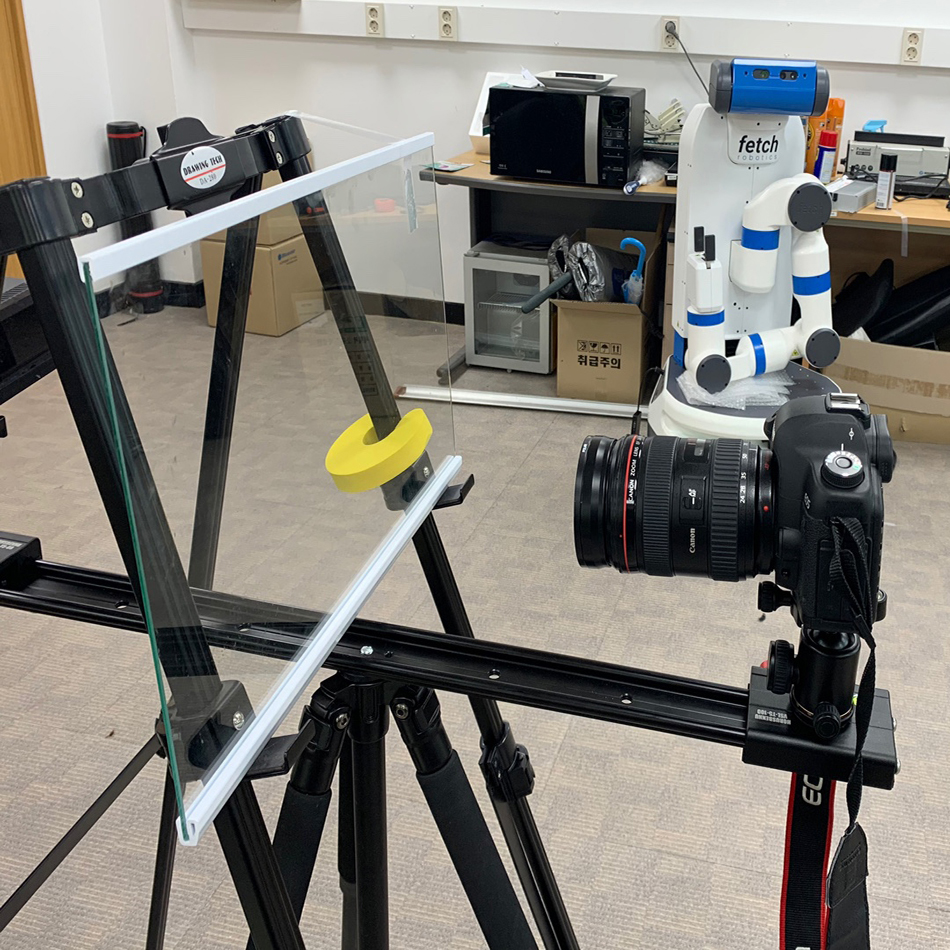}
		\caption*{\footnotesize{With glass}}
	\end{subfigure}\hspace{1mm}
	\begin{subfigure}{.11\textwidth}
		\centering
		\includegraphics[width=\linewidth]{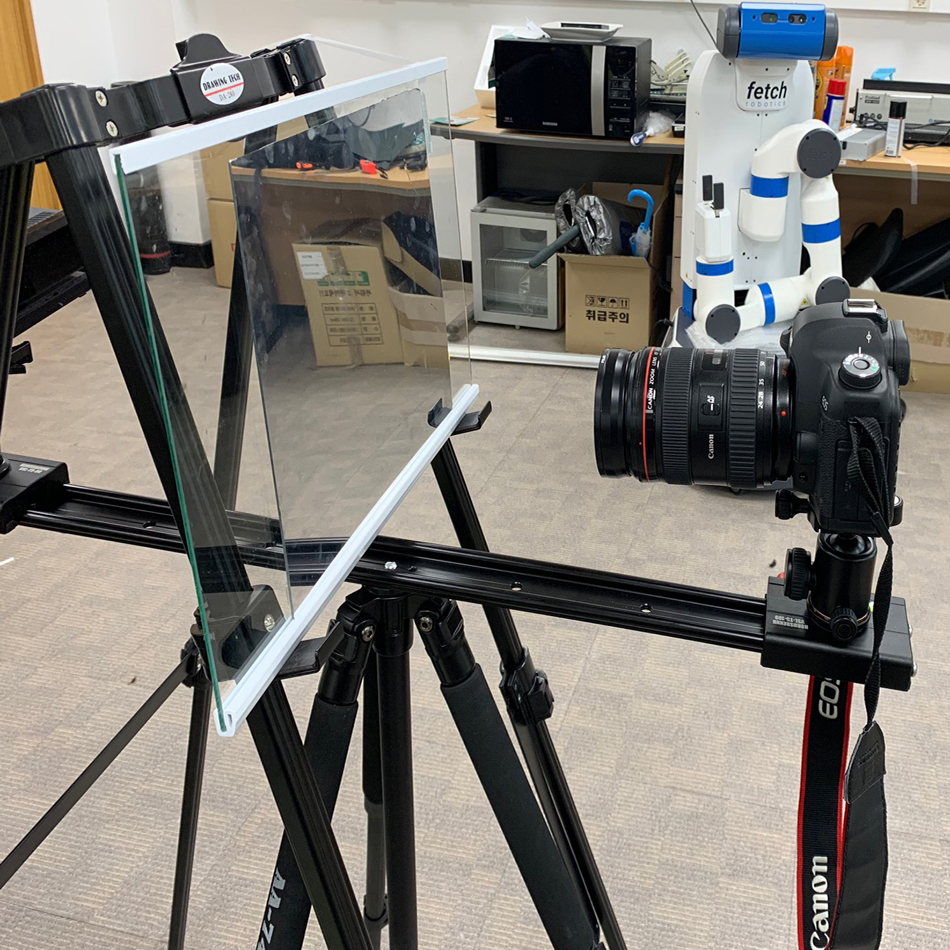}
		\caption*{\footnotesize{With mirror}}
	\end{subfigure}\hspace{1mm}
	\begin{subfigure}{.11\textwidth}
		\centering
		\includegraphics[width=\linewidth]{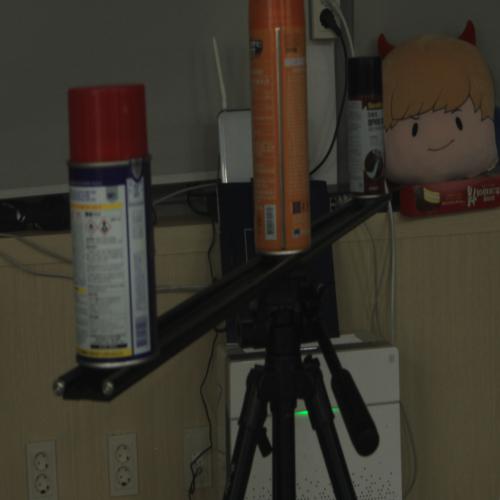}
		\caption*{\footnotesize{Front scene}}
	\end{subfigure}\hspace{1mm}
	\begin{subfigure}{.11\textwidth}
		\centering
		\includegraphics[width=\linewidth]{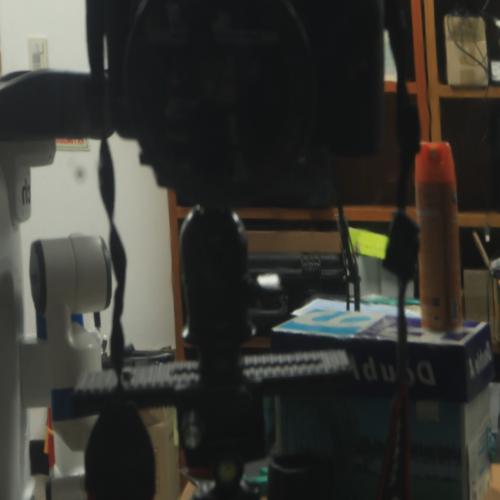}
		\caption*{\footnotesize{Back scene}}
	\end{subfigure}
	\vspace{-3mm}
	\caption{
		\small{Experiment setup and taken images. 
		}
	}
	\label{fig:experiment_setup}
\end{figure}

\begin{figure*}[h]
	\begin{center}
	\includegraphics[width=\linewidth]{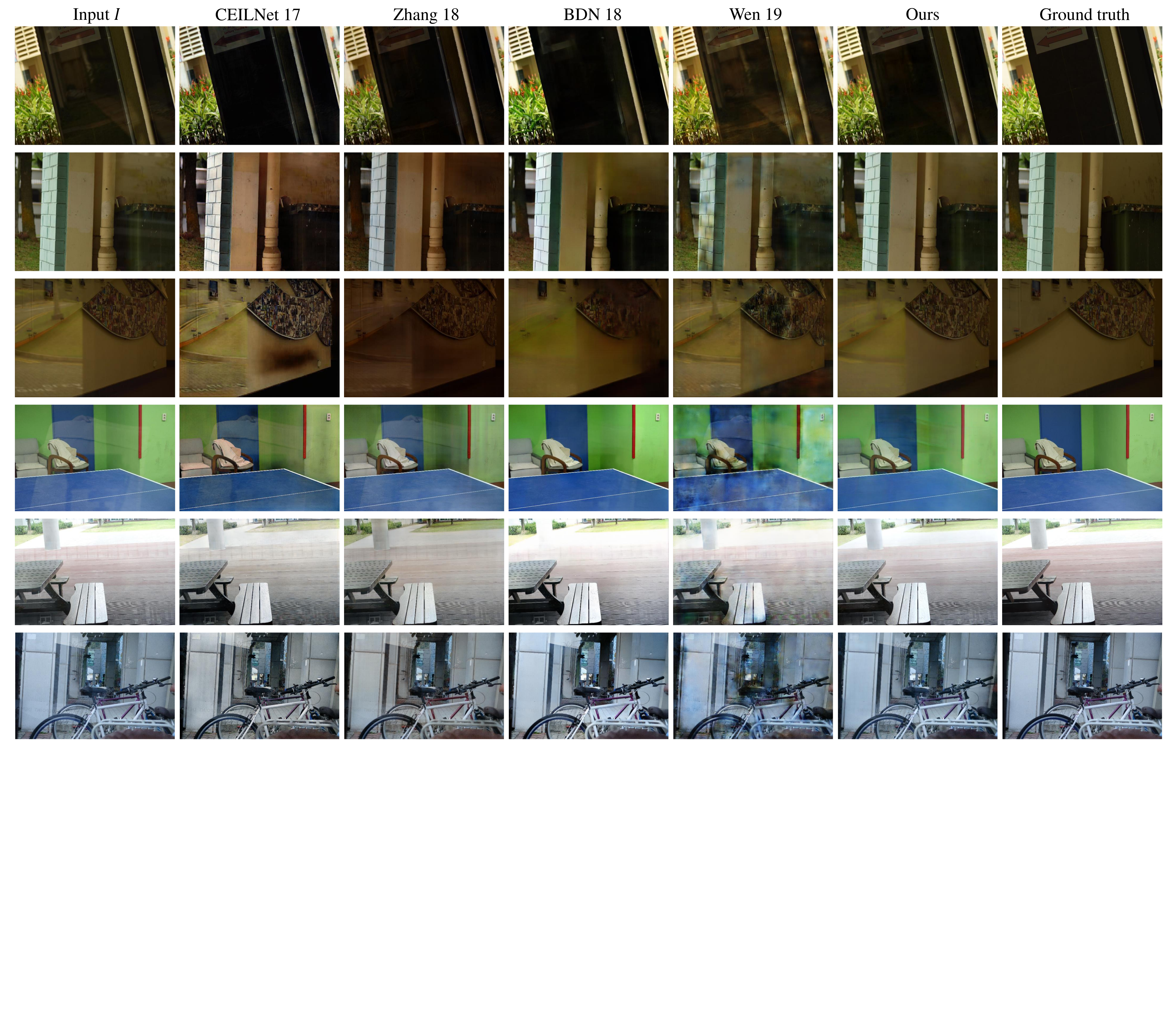}
	\end{center}
\vspace{-3mm}
	\caption[]{\label{fig:wild}
		\small{
			Examples of reflection removal results on the wild 
dataset (Rows 1-3) and our real 100 testset (Rows 4-6) visually. 	
		} 	
	}	

\end{figure*}

\Skip{
\begin{figure*}[h]
	\begin{center}
		\includegraphics[width=0.9\linewidth]{images/rendered_results.pdf}
	\end{center}

	\caption[]{\label{fig:rendered}
		\small{
		Reflection removal example on rendered test set.
		} 	
	}	

\end{figure*}
}

\paragraph{Implementation.}
Each of our two sub-nets shares the same structure based on the one proposed in
\cite{zhang2018single}, and they are fully
convolutional networks for considering global information.
For the training, we first train
$BT$-net with rendered image pairs independently, and then
pre-trained $BT$-net is connected to $SP$-net for $SP$-net training ($BT$-net is fine-tuned in this stage). $SP$-net is trained by minimizing the aforementioned
loss terms between GTs and their predictions with a learning rate of
$10^{-4}$. The rendered training images have the 256 $\times$ 256 resolution. 

\section{Experiments with Real and Synthetic Data}
\label{sec:Experiments}

We compare our approach with the state-of-the-art deep learning-based
reflection removal methods,
CEILNet~\cite{fan2017generic},
Zhang et al.~\cite{zhang2018single}, BDN~\cite{yang2018seeing}, Wen et al.~\cite{wen2019single}
across different test sets that work for a given
single image.

For quantitative evaluation with real-world images, we utilize the well-known
reflection removal benchmark, the SIR Wild dataset~\cite{wan2017benchmarking}. It
consists of three images ($I, T,\tilde{R}$) under various capturing settings from controlled indoor scenes to wild scenes. Since the indoor dataset is designed for exploring the impact of
various parameters, we test our results on their wild scenes. Also, we additionally capture 100 real reflection pair images for testing (denoted as real100). Also, we generate 200 rendered images for testing.

\begin{table}[b]
	\small
	\centering
	\begin{tabular}{l|l|l}
		\hline
		Method         & PSNR   & SSIM   \\ \hline
		CEIL \cite{fan2017generic}   & 14.466 & 0.737  \\ \hline
		Zhang \cite{zhang2018single} & 20.379 & 0.842 \\ \hline
		Wen \cite{wen2019single}   & 20.266 & 0.856  \\ \hline
		Ours           & 29.307 & 0.943  \\ \hline
	\end{tabular}
	\vspace{-2mm}	
	\caption{\small{The average similarity of synthesized reflections with 10 real camera-captured reflection images 
	}}
	\label{tb2:dataAnalysis}
\end{table}



\begin{table*}[]
	\small
	\centering
	\begin{tabular}{|c|c|c|c|c|c|r|c|c|c|c|c|}
		\hline
		\multirow{2}{*}{Dataset}   & \multirow{2}{*}{Index} & \multicolumn{10}{c|}{Methods}                                                                                                                                                                                                                 \\ \cline{3-12} 
		&          & \multicolumn{1}{c|}{Input} & \multicolumn{1}{c|}{\begin{tabular}[c]{@{}c@{}}CEILNet\\ \cite{fan2017generic}
		\end{tabular}}& \multicolumn{1}{c|}{\begin{tabular}[c]{@{}c@{}}CEILNet\\ FW\end{tabular}} & \multicolumn{1}{c|}{\begin{tabular}[c]{@{}c@{}}CEILNet\\ FR\end{tabular}} & \multicolumn{1}{c|}{\begin{tabular}[c]{@{}c@{}}Zhang\\ \cite{zhang2018single}\end{tabular}} & \multicolumn{1}{c|}{\begin{tabular}[c]{@{}c@{}}Zhang\\ FW\end{tabular}} & \multicolumn{1}{c|}{\begin{tabular}[c]{@{}c@{}}Zhang\\ FR\end{tabular}} & \multicolumn{1}{c|}{\begin{tabular}[c]{@{}c@{}}BDN\\ \cite{yang2018seeing}\end{tabular}} & \multicolumn{1}{c|}{\begin{tabular}[c]{@{}c@{}}Wen\\ \cite{wen2019single}\end{tabular}} & \multicolumn{1}{c|}{Ours} \\ \hline
		\multirow{2}{*}{SIR Wild \cite{wan2017benchmarking}}  
		& PSNR &  \textcolor{red}{25.89}   & 20.89 &  19.23    &    22.51  & 21.15     &21.34    &  23.18  & 22.02    &   21.26   &   \textcolor{blue}{25.55}     \\ \cline{2-12} 
		& SSIM    & \textcolor{blue}{0.903}  & 0.826 &  0.819   &  0.880      &   0.851  & 0.865      &   0.890     &   0.835  & 0.835   &    \textcolor{red}{0.905}       \\ \hline
		\multirow{2}{*}{Real 100}                                                   
		& PSNR   &  \textcolor{blue}{21.53}   &  19.24 &   17.82  & 20.35  & 18.66 &  18.88   & 20.44    &    19.46    &   19.07 & \textcolor{red}{21.59} \\ \cline{2-12} 
		& SSIM   & \textcolor{red}{0.797}  &  0.733 &   0.706  &  0.764     & 0.750 & 0.753       &     0.773  &  0.753      & 0.728 &   \textcolor{blue}{0.789} \\ \hline
		\multirow{2}{*}{\begin{tabular}[c]{@{}c@{}}Rendered\\ Testset\end{tabular}} 
		& PSNR  & 23.27   & 19.31  &   20.23 &  23.46&  22.21 &  21.83  &  \textcolor{blue}{24.43}    & 21.66 & 21.79 &   \textcolor{red}{27.90}                 \\ \cline{2-12} 
		& SSIM  &0.846&0.745& 0.777 &  0.829  &    0.829  & 0.828 &   \textcolor{blue}{0.854}  & 0.819 &  {0.804} & \textcolor{red}{0.894}                 \\ \hline
	\end{tabular}
	\caption{
		\small{
			Quantitative results of different methods on SIR wild, our real 100, and rendered test set. 
			Some result images of the SIR dataset can be found in
			Figure~\ref{fig:wild}. CEILNet, Zhang, and BDN are the
			pre-trained networks. CEILNet-FR and Zhang-FR are fine-tuned with our rendered training images, and CEILNet-FW and Zhang-FW are fine-tuned with Wen's data generation method with the same source images with ours. Red numbers are the \textcolor{red}{best}, and blue numbers are the \textcolor{blue}{second best} results.}
	}
\vspace{-3mm}
\label{tb2:quantitative}
\end{table*}
\subsection{Dataset Evaluation}
\label{sec:DataAnalysis}

 In order to validate our rendered dataset and its similarity to real-world reflection captured by a camera, we capture
real image pairs with devices of Figure~\ref{fig:experiment_setup}. We first
capture the GT $I$ with a glass (so that it contains reflection), then use a mirror to capture GT $R$ and remove
the glass to capture GT $T$ as inputs of data synthesis. In order to match
common RGB and RGBD datasets, the GT $T$ and $R$ are captured with F22 to
minimize the defocusing effect. In addition, we capture and calibrate the depth map
using a Kinect on each side of the slider across the glass. With the captured GT $T$
and $R$, we generate reflection images with three different
methods~\cite{fan2017generic,zhang2018single,wen2019single}, and compare them
with our rendered images. Figure~\ref{fig:datacomp} shows an example of the generated
reflections. As shown, with depth information and a physically based rendering model, ours can generate lens- and glass-effects much similar to the real images.
 
Table~\ref{tb2:dataAnalysis} shows the numerical comparison of generated
reflection images, and we use average PSNR and SSIM for measuring the similarity. We take two different scenes with 5 focus points, in total
10 real reflection images for comparison. Note that 10 real reflection images are different from the real 100 test set we captured because the real 100 test set does not have depth. For a fair comparison, we randomly synthesize 100 images using both CEIL~\cite{fan2017generic} method and Zhang et al.~\cite{zhang2018single} method for every 10 scenes and pick the best PSNR and SSIM synthesized image for each scenes. For the Wen~\cite{wen2019single} method, since their method utilizes pre-trained reflection synthesis network to produce 3 types of reflection, we generate 3 different images for each scene. Among them, we pick the best PSNR and SSIM synthesized result for each scenes. The report of their average values is listed in Table~\ref{tb2:dataAnalysis}.


\subsection{Ablation Study}
In order to validate the effectiveness of a priori loss from the $BT$-net, we
evaluate each model (w/ and w/o a priori loss) on SIR wild, real 100 images,
and 200 rendered images. Each model is trained from scratch with a denoted
loss combination. 
Since we followed the other loss terms of Zhang et
al.~\cite{zhang2018single}, we conduct an ablation study on the new a
priori loss only.

\begin{table}[t]
	\small
	\begin{tabular}{|c|c|c|c|}
		\hline
		\multirow{2}{*}{Dataset}        & \multirow{2}{*}{Index} & \multicolumn{2}{c|}{Model}                                                            \\ \cline{3-4} 
		&                        & w/o a priori loss & \begin{tabular}[c]{@{}c@{}} w/ a priori loss\\ (Ours)\end{tabular} \\ \hline
		\multirow{2}{*}{SIR wild \cite{wan2017benchmarking}}     & PSNR                   & 24.31             & \textbf{25.54}                                                    \\ \cline{2-4} 
		& SSIM                   & 0.874             & \textbf{0.905}                                                    \\ \hline
		\multirow{2}{*}{Real 100}                                                   & PSNR                   & 20.86             & \textbf{21.58}                                                    \\ \cline{2-4} 
		& SSIM                   & 0.772             & \textbf{0.789}                                                    \\ \hline
		\multirow{2}{*}{\begin{tabular}[c]{@{}c@{}}Rendered\\ Testset\end{tabular}} & PSNR                   & 27.34             & \textbf{27.90}                                                    \\ \cline{2-4} 
		& SSIM                   & 0.889             & \textbf{0.894}                                                    \\ \hline
	\end{tabular}
	\caption{\small{Quantitative comparison of our ablated models}}
	\label{tb1:ablation}
	\vspace{-4mm}
\end{table}

\begin{figure}[h]
	\begin{center}
		\includegraphics[width=1\linewidth]{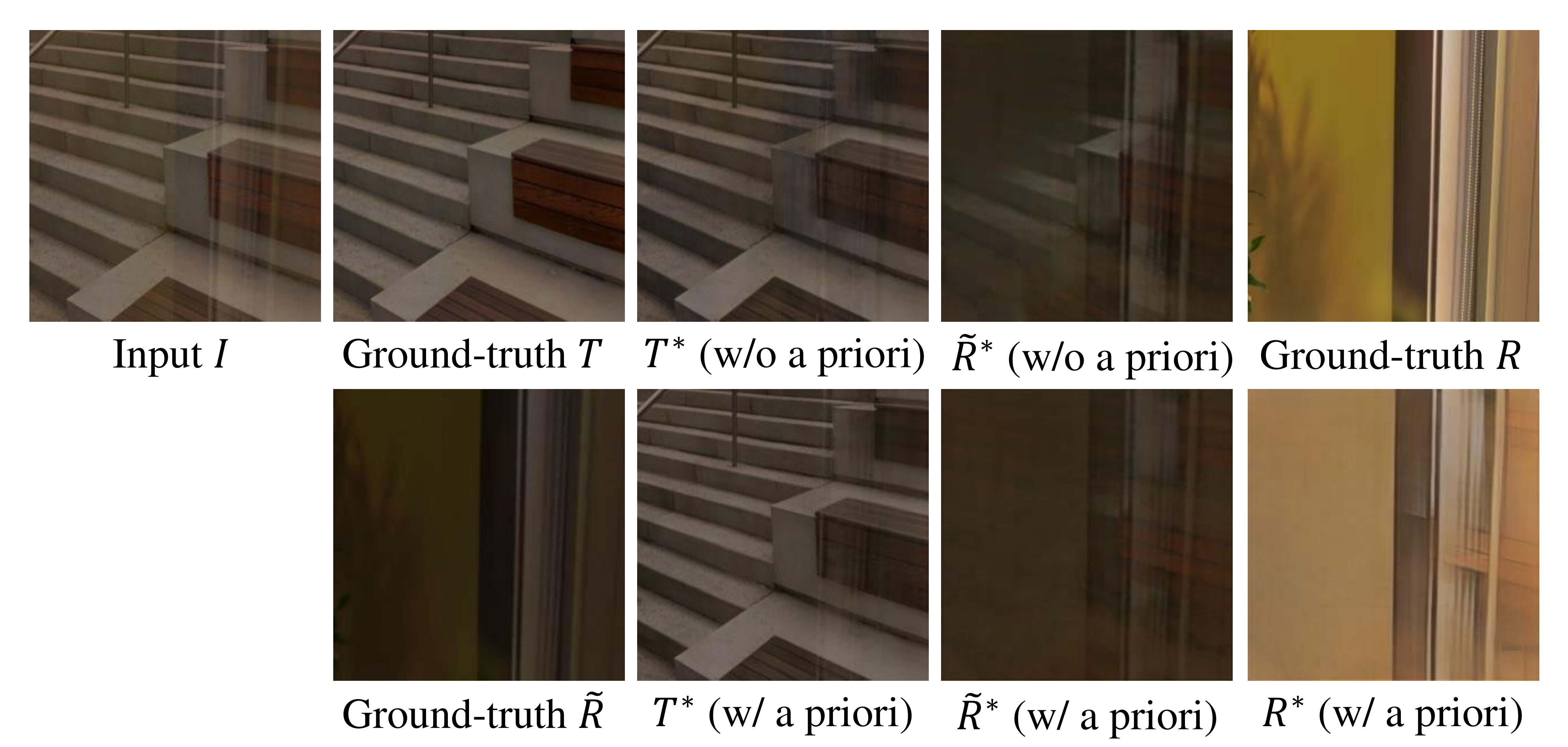}
	\end{center}
\vspace{-4mm}
	\caption[]{\label{fig:ablation}
		\small{Comparison between the rendered results of our complete model and an ablated model.} 
	}	
\vspace{-3mm}
\end{figure}

The numerical results show that using the additional a priori loss can improve
the separation quality both in the real and rendered test sets. Since
$BT$-net backtracks the darken and distorted predicted $\tilde{R}^*$ into $R^*$
to calculate a priori loss, this loss can provide a more robust signal of
separation quality. Moreover, Figure~\ref{fig:ablation} shows some visual
results of our complete model and ablated model on the rendered test set. Since
$BT$-net can backtrack the predicted $\tilde{R}^*$ into  ${R}^*$, our complete
model could figure out the reflection and transmission area better when
separating.

\subsection{Comparision on Benchmarks}

For comparison, we utilize pre-trained network weights provided by authors.
Additionally, we also fine-tune the author's pre-trained network with our rendered
dataset and a dataset generated by Wen's reflection synthesis network. Two
generated datasets share the same source image pairs, and we use Wen's pre-trained
weight and default setting for generating reflection training images. Both
fine-tuned networks are tuned with the same epoch and learning rate. We name the
models that are fine-tuned with our rendered data with a suffix `$-FR$', and the
models finetuned with Wen's reflection synthesized images with a suffix
`$-FW$'. Since BDN does not provide training code, and Wen's network needs
additional alpha blending mask for training their separation network, we cannot fine-tune them.

Figure~\ref{fig:wild} shows some visual examples of reflection removal results
on the SIR wild test set and our real 100 test set. All the compared methods do not
work well in terms of removing strong regional reflection (row 3), but still,
our method removes some of the reflection without significantly damaging the
transmission area. In the last row, ours and BDN~\cite{yang2018seeing} could
remove the reflection of banner in the below, while other methods do not remove,
but darken the overall intensity.

Table~\ref{tb2:quantitative} shows quantitative results on the
real-world test sets (SIR wild and real 100) and our rendered test set. We utilize SSIM and PSNR as error metrics, which are widely used in prior reflection removal methods. 
Our method achieves the best or second-best numerical results in all the
datasets. We also validate that our dataset can improve previous methods (pre-trained
networks) by supplying more physically-based reflection training images
(Table~\ref{tb2:quantitative}).
However, for both real reflection testset, none of the existing methods, no matter
how they are trained by classic synthesized dataset or our rendered dataset,
outperforms the unseparated input in both error metrics. This suggests there is
still room for further improvement. 
\vspace{-1mm}

\section{Conclusion}
\label{sec:conclusion}

We have proposed a novel learning-based single image reflection removal method,
which utilizes reflection training images generated by physically-based
rendering. The training images consist of different types, including transmission and
reflection w/ and w/o the glass/lens-effects, and provide both classical a posteriori
and novel a priori information. With the new dataset, we proposed $SP$-net to separate the input into two layers with the help of $BT$-net to remove the glass/lens-effect in the separated layers for error calculation (a priori loss). With a priori loss, the separation loss calculation is improved. Also, we validated that our physically-based training data can improve existing learning-based reflection removal methods as well with various real reflection test images.

\paragraphTitle{Limitation.} 
 In this paper, we did not consider viewpoints that are not perpendicular to the glass. That is one possible extension for future research. Also, we did not consider the curved glass or glass with a special shape, while our rendering approach can accommodate these cases by replacing the plane glass model with a curved one in the future. 
\Skip{While our method works well with a wide type of real images. 
It has a certain limitation.

In an extreme case, there can be a strong reflection point if there
is a light in the background ($R$) with relatively dark foreground ($T$). 
Since our training dataset does not contain those case, our method cannot recognize those parts as reflection and fails to
remove it as shown in Figure~\ref{fig:limitation}.

}
 \vspace{-1mm}

\section*{Acknowledgments}
 \vspace{-1mm}
We would like to thank anonymous reviewers for constructive comments. 
Sung-Eui Yoon and Yuchi Huo are co-corresponding authors of the paper. 
This work was supported by MSIT/NRF (No. 2019R1A2C3002833) and SW Starlab program (IITP-2015-0-00199)

{\small
\bibliographystyle{ieee_fullname}
\bibliography{egbib}
}

\end{document}